\documentclass[sigconf]{acmart}
\renewcommand\footnotetextcopyrightpermission[1]{}
\setcopyright{none}
\acmConference[]{}{}{}     
\acmBooktitle{}
\acmPrice{}
\acmISBN{}
\acmDOI{}

\usepackage{booktabs}   
\usepackage{tabularx}   
\usepackage{graphicx}  %

\AtBeginDocument{%
  }

\setcopyright{acmlicensed}
\copyrightyear{2025}
\acmYear{2025}
\acmDOI{XXXXXXX.XXXXXXX}
\acmISBN{978-1-4503-XXXX-X/2018/06}

\begin{document}

\title{Abundance-Aware Set Transformer for Microbiome Sample Embedding}

\author{Hyunwoo Yoo}
\email{hty23@drexel.edu}
\affiliation{%
  \institution{Drexel University}
  \city{Philadelphia}
  \state{Pennsylvania}
  \country{USA}
}

\author{Gail Rosen}
\email{glr26@drexel.edu}
\affiliation{%
  \institution{Drexel University}
  \city{Philadelphia}
  \state{Pennsylvania}
  \country{USA}
}


\begin{abstract}
Microbiome sample representation to input into LLMs is essential for downstream tasks such as phenotype prediction and environmental classification. While prior studies have explored embedding-based representations of each microbiome sample, most rely on simple averaging over sequence embeddings, often overlooking the biological importance of taxa abundance. In this work, we propose an abundance-aware variant of the Set Transformer to construct fixed-size sample-level embeddings by weighting sequence embeddings according to their relative abundance. Without modifying the model architecture, we replicate embedding vectors proportional to their abundance and apply self-attention-based aggregation. Our method outperforms average pooling and unweighted Set Transformers on real-world microbiome classification tasks, achieving perfect performance in some cases. These results demonstrate the utility of abundance-aware aggregation for robust and biologically informed microbiome representation. To the best of our knowledge, this is one of the first approaches to integrate sequence-level abundance into Transformer-based sample embeddings.
\end{abstract}




\maketitle

\pagestyle{empty}    
\fancyhead{}         
\fancyfoot{}         
\thispagestyle{empty} 

\section{Introduction}

Microbiome samples contain thousands of short DNA sequences derived from diverse microbial species in an environmental or host-associated context~\cite{mendler2024microbiome}. A central challenge in microbiome analysis is to transform such variable-length, unordered sets of sequences into fixed-size, informative representations that can support downstream tasks such as phenotype prediction, environmental classification, or disease detection. These sample-level embeddings are not only critical for training classifiers such as deep neural networks or random forests, but also serve as the basis for similarity comparisons, clustering, and visualization.

Recent approaches typically employ pretrained DNA language models such as DNABERT~\cite{zhou2023dnabert} or the Nucleotide Transformer~\cite{dalla-torre2023nucleotide} to compute embeddings for individual sequences. These sequence embeddings are then aggregated—usually via mean or max pooling—to obtain a single vector per sample~\cite{zaheer2017deep}. While simple and computationally efficient, these aggregation strategies fail to account for key biological signals such as the abundance of each 16S rRNA and/or amplicon  sequence  variant  (ASV), which reflects the relative prevalence of microbial species in the sample~\cite{oh2020deepmicro, mandal2015analysis}.

Abundance is not merely a statistical measure; it often carries meaningful biological information. For instance, the abundance of certain taxa may indicate infection status, environmental perturbation, or ecological balance. In disease-related microbiome studies, rare pathogenic organisms may increase in abundance over time, while in ecological monitoring, the presence and concentration of specific indicator taxa can signal pollution levels or habitat changes. In Zhao et al. \cite{10.1371/journal.pcbi.1009345}, 3 types of sample classification are considered: 1) reads are classified to sample attributes, and the overall sample attribute is the final phenotype classification; 2) the sample-level embedding method forms a sample-level vector representation by averaging all read-level embeddings in a query sample; and 3) for the Pseudo OTU method, as described by Woloszynek et al. \cite{10.1371/journal.pcbi.1006721}, reads are embedded as vectors and are clustered into Pseudo OTUs (groupings of related read vectors). Then, each query sample’s reads were assigned to those Pseudo OTUs based on distance. The latter method (c) ended up having the best performance, since it took into account the abundance of taxa the best. Simple averaging and the majority vote obscured taxa-abundance, which are important signals, especially when low-abundance but high-importance sequences are diluted by a large number of irrelevant ones.

To address this limitation, we propose an {\bf abundance-aware Set Transformer} architecture for constructing sample embeddings. The Set Transformer~\cite{lee2019set} is a permutation-invariant attention-based model that has shown promise in learning representations from unordered sets. Our proposed method integrates abundance information through two strategies: (1) {\bf repetition-based weighting}, where each sequence embedding is repeated in proportion to its abundance, and (2) {\bf soft attention weighting}, where abundance is incorporated directly into the pooling weights. These methods require no architectural changes and can be easily implemented with existing attention-based modules.

\begin{figure*}[h!]
    \centering
    \includegraphics[width=1\linewidth]{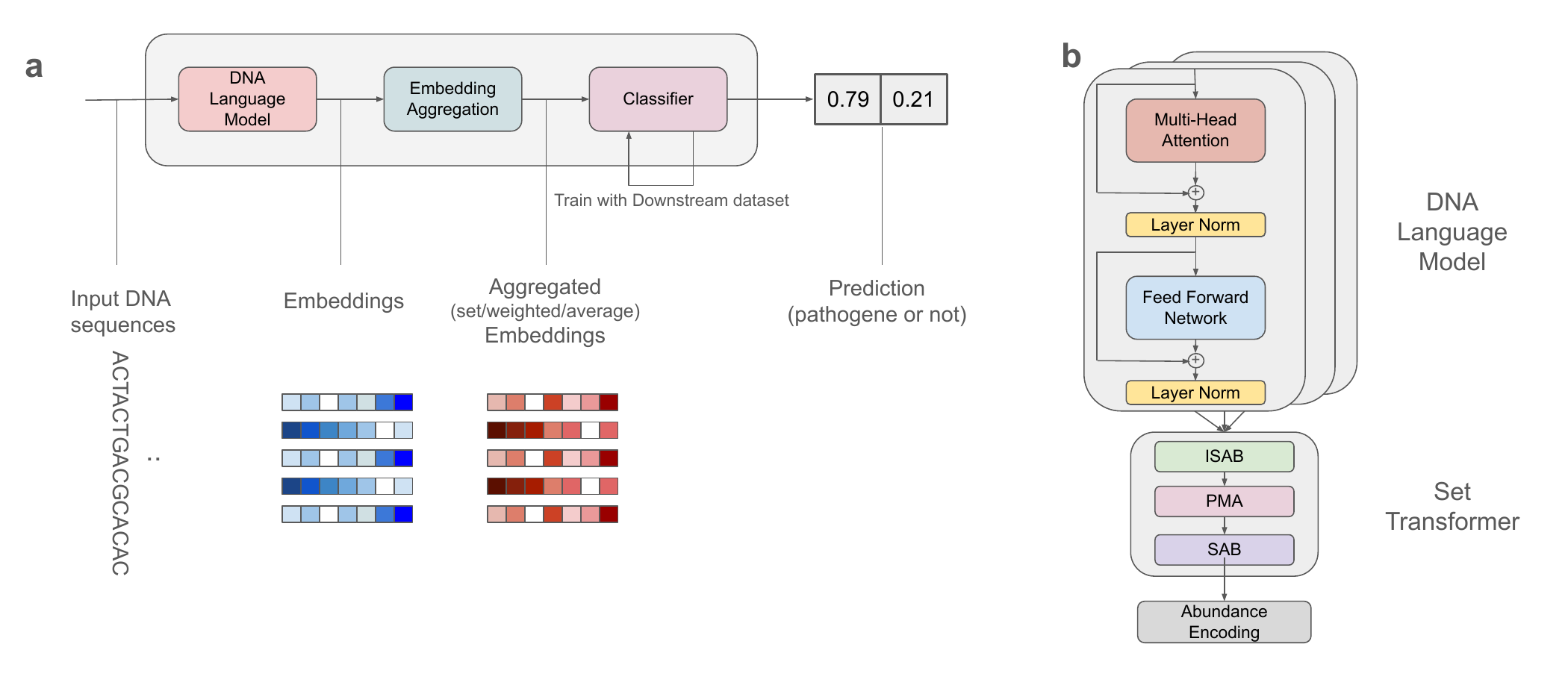}
    \caption{Overview of the abundance-aware set transformer architecture.
(a) End-to-end predictive pipeline in which input DNA sequences are first encoded into contextual embeddings by a DNA language model, then aggregated by an abundance-aware set transformer to produce a fixed-size, weighted representation, and finally passed to a downstream classifier that outputs class probabilities (e.g. pathogen or not). The classifier (a fully-connected neural network or random forest) is trained on a labeled downstream dataset.
(b) Abundance-aware Set Transformer architecture, where a DNA language model encoder (stack of multi-head attention, feed-forward networks, and layer normalization) first encodes each sequence into a contextual representation, followed by a Set Transformer comprising ISAB, PMA, and SAB blocks to aggregate sequence embeddings into a fixed-size set representation, and an abundance encoding mechanism that weights the set elements according to their relative sequence abundance before pooling. Specifically, ISAB (Induced Set Attention Block) efficiently reduces computational complexity by introducing a small number of inducing points to capture set-wise interactions, PMA (Pooling by Multihead Attention) uses learnable seed vectors to produce a fixed-size embedding that summarizes the entire set, and SAB (Set Attention Block) refines these representations by modeling self-attention among the pooled set embeddings to capture higher-order relationships.}
    \label{fig:overview}
\end{figure*}

Unlike traditional models that treat DNA sequences independently, our approach captures both the content and {\bf structure} of the full microbiome sample by leveraging abundance during the aggregation step. This allows the model to focus on biologically meaningful signals, improving both performance and interpretability.

We evaluate our method across three real-world microbiome classification tasks involving both clinical and environmental samples. Our results show that abundance-aware Set Transformers consistently outperform standard mean pooling and vanilla Set Transformers. In some cases, our model even achieves perfect classification accuracy. These findings underscore the importance of integrating abundance as a first-class signal in set-based representations for microbiome analysis.

\paragraph{Our contributions are summarized as follows:}
\begin{itemize}
    \item We propose an abundance-aware Set Transformer to generate biologically meaningful sample-level embeddings from microbiome sequencing data.
    \item We compare four aggregation strategies—mean pooling, soft weighting, Set Transformer, and abundance-aware Set Transformer—across three distinct classification tasks.
    \item We demonstrate that incorporating abundance significantly improves model performance, especially in challenging cross-study settings, without requiring modifications to the base Transformer architecture.
\end{itemize}

\section{Related Work}

\subsection{Pretrained DNA language models.}

Large-scale language models trained on genomic sequences have significantly advanced sequence-level representation learning. 

DNABERT~\cite{Ji2021DNABERT, zhou2023dnabert} and Nucleotide Transformer~\cite{dalla-torre2023nucleotide, mendoza2024foundational} employ transformer based architectures~\cite{vaswani2017attention} to embed k-mer tokenized DNA sequences, and have demonstrated strong performance in tasks such as promoter prediction, splice site detection, and enhancer identification. However, these models operate at the level of individual sequences and do not directly address the challenge of aggregating thousands of such embeddings into a coherent sample-level representation.

\begin{table*}[h!]
\centering
\caption{FCNN classification and Random Forest results on Study 14245.}
\label{tab:results_14245}
\begin{tabular}{lccccc}
\toprule
\textbf{Embedding Method} & \textbf{Classifier} &\textbf{Accuracy} & \textbf{Macro Precision} & \textbf{Macro Recall} & \textbf{Macro F1} \\
\midrule
Set Transformer           &FCNN & 0.5417 & 0.2708 & 0.5000 & 0.3514 \\
Average                   &FCNN & 0.5417 & 0.2708 & 0.5000 & 0.3514 \\
Weighted Average          &FCNN & 0.5417 & 0.2708 & 0.5000 & 0.3514 \\
Weighted Set Transformer  &FCNN & 0.5417 & 0.2708 & 0.5000 & 0.3514 \\
\midrule
Set Transformer           &Random Forest & 0.5417 & 0.5313 & 0.5280 & 0.5209 \\
Average                   &Random Forest & 0.5000 & 0.4889 & 0.4895 & 0.4857 \\
Weighted Average          &Random Forest & 0.5417 & 0.5417 & 0.5420 & 0.5409 \\
Weighted Set Transformer  &Random Forest & \textbf{0.5833} & \textbf{0.5804} & \textbf{0.5804} & \textbf{0.5804} \\
\bottomrule
\end{tabular}
\end{table*}

\subsection{Sample-level representation learning.}
To derive a fixed-size vector for a microbiome sample, a common strategy is to apply simple pooling operations such as mean or max pooling over all sequence embeddings~\cite{lo2019metann, arumugam2011enterotypes}. While computationally efficient, these operations treat all sequences equally, ignoring important biological signals like taxonomic relevance or abundance. Recent studies in other domains have explored more expressive aggregation methods using attention mechanisms. The Set Transformer~\cite{lee2019set}, for example, enables permutation-invariant learning over sets via attention-based pooling and has shown success in tasks such as point cloud classification and multi-instance learning. In microbiome research, its use has been limited and typically does not incorporate abundance metadata~\cite{martino2019novel, sankaran2019latent}. Recent work such as SetBERT~\cite{Ludwig2025} extends the Set Transformer framework to high-throughput sequencing data, enabling contextualized sample-level embeddings and explainable predictions, though it does not explicitly incorporate abundance information into the aggregation process.

\subsection{Incorporating abundance in microbiome modeling.}
Abundance information plays a critical role in microbiome analysis, reflecting the prevalence of different taxa in a sample. Traditional bioinformatics pipelines (e.g., QIIME2~\cite{bolyen2019qiime2}) and microbiome-specific machine learning tools like DeepMicro~\cite{oh2020deepmicro}, METAML~\cite{pasolli2016mlmeta}, and MicroKPNN-MT~\cite{monshizadeh2025multitask} integrates taxonomic abundance and clinical metadata using a multitask neural network for phenotype prediction. However, these methods usually depend on OTU/ASV-level tables or taxonomic annotations, and do not operate directly on raw sequencing data. To our knowledge, no prior work has combined abundance information with DNA-level language model embeddings within a set-based deep learning framework.

Our work fills this gap by integrating abundance-aware attention into Set Transformer-based aggregation of DNA language model embeddings, enabling sample-level representation learning that reflects both sequence content and quantitative structure.

\section{Methods}

We propose a pipeline for constructing abundance-aware microbiome sample embeddings and evaluating their utility in downstream classification. As illustrated in Figure~\ref{fig:overview}, the methodology consists of four main stages including sequence embedding, sample-level embedding aggregation using a Set Transformer architecture, dataset construction, and classification. We describe each component in detail below, along with a deeper explanation of the architectural components shown in Figure~\ref{fig:overview}(a) and (b).

\subsection{Sequence Embedding with DNABERT-2}

Each input microbiome sample consists of a set of unique nucleotide sequences with associated abundance values. We utilized the pretrained DNABERT-2 model~\cite{zhou2023dnabert} to obtain contextual embeddings for each sequence. DNABERT-2 is a Transformer-based DNA language model composed of stacked encoder blocks, each consisting of multi-head self-attention layers, position-wise feedforward networks, and layer normalization.

Each nucleotide sequence was first tokenized into DNABERT-2 tokens and fed into the model. We extracted the embedding corresponding to the [CLS] token from the final encoder layer, yielding a 768-dimensional vector representation for each sequence. Only sequences with non-zero abundance were retained for downstream processing.

\subsection{Sample-Level Embedding Aggregation}

To construct a fixed-size vector representation for each microbiome sample, we employed four aggregation strategies applied to the sequence-level embeddings including average pooling~\cite{mikolov2013efficient}, weighted average pooling~\cite{arora2017simple}, set transformer aggregation~\cite{lee2019set}, and weighted set transformer:

All strategies yield a 768-dimensional embedding \( \mathbf{z}_S \in \mathbb{R}^{768} \) for each sample \( S \).

  \textbf{Average Pooling}: The mean of all unique sequence embeddings associated with the sample, ignoring abundance. This assumes uniform contribution across sequences:
  \[
  \mathbf{z}_S = \frac{1}{N} \sum_{i=1}^N \mathbf{e}_i
  \]

  \textbf{Weighted Average Pooling}: Each embedding is weighted by its relative abundance \( a_i \), normalized across the sample:
  \[
  \mathbf{z}_S = \sum_{i=1}^N \alpha_i \mathbf{e}_i, \quad \text{where } \alpha_i = \frac{a_i}{\sum_{j=1}^N a_j}
  \]

  \textbf{Set Transformer Aggregation}: The full set of sequence embeddings (including repetitions according to observed abundance) is passed to a Set Transformer~\cite{lee2019set}. This allows implicit modeling of abundance via sequence duplication without architectural modification.

\begin{table*}[h!]
\centering
\caption{FCNN and Random Forest classification results on Study 10442.}
\label{tab:results_10442}
\begin{tabular}{lccccc}
\toprule
\textbf{Embedding Method} & \textbf{Classifier} & \textbf{Accuracy} & \textbf{Macro Precision} & \textbf{Macro Recall} & \textbf{Macro F1} \\
\midrule
Set Transformer           &FCNN & 0.9200 & 0.4600 & 0.5000 & 0.4792 \\
Average                   &FCNN & 0.9200 & 0.4600 & 0.5000 & 0.4792 \\
Weighted Average          &FCNN & 0.9200 & 0.4600 & 0.5000 & 0.4792 \\
Weighted Set Transformer  &FCNN & \textbf{1.0000} & \textbf{1.0000} & \textbf{1.0000} & \textbf{1.0000} \\
\midrule
Set Transformer           &Random Forest & 0.9533 & 0.9759 & 0.7083 & 0.7818 \\
Average                   &Random Forest & 0.9533 & 0.8712 & 0.7844 & 0.8208 \\
Weighted Average          &Random Forest & 0.9600 & 0.9792 & 0.7500 & 0.8227 \\
Weighted Set Transformer  &Random Forest & \textbf{1.0000} & \textbf{1.0000} & \textbf{1.0000} & \textbf{1.0000} \\
\bottomrule
\end{tabular}
\end{table*}

  \textbf{Weighted Set Transformer (Ours)}: Unique sequence embeddings are passed to the Set Transformer once, and abundance is incorporated post-aggregation via a soft weighting scheme over the output vectors. This strategy enables integration of abundance while avoiding sequence repetition.

\subsection{Architecture of Weighted Set Transformer Pipeline} 
As shown in Figure~\ref{fig:overview}(b), our Set Transformer Pipeline module consists of the following components:

  \textbf{ISAB (Induced Set Attention Block)} reduces the quadratic complexity of self-attention by introducing a small number of learned inducing points. These inducing points summarize the interactions within the set and serve as a bottleneck for efficient attention computation.

  \textbf{PMA (Pooling by Multihead Attention)} uses learnable seed vectors to pool the set into a fixed-size embedding. Each seed attends over the set elements, forming a summary representation.

  \textbf{SAB (Set Attention Block)} applies self-attention among the pooled outputs to refine and model higher-order dependencies between them. 
  Each of these components is permutation-invariant, allowing the model to robustly aggregate unordered sets of DNA sequence embeddings.

  \textbf{Abundance-Aware Pooling.} As a final step, we integrate abundance information into the output of the Set Transformer. Rather than duplicating sequence embeddings, we apply abundance-aware weights to the output vectors. Specifically, we compute:
\[
\mathbf{z}_S = \sum_{i=1}^{N} \alpha_i \mathbf{o}_i
\]
where \( \mathbf{o}_i \) are the output embeddings from the Set Transformer and \( \alpha_i \) are abundance-normalized weights.

This soft-weighting approach preserves the biological signal of relative abundance while avoiding computational overhead from repetition.



  

  



\subsection{Labeling and Dataset Construction}

Sample-level metadata were obtained from the Qiita portal and labeled according to the \texttt{primary\_experimental\_variable} field. Samples labeled as \textit{tumor mucosa} were assigned label 1, and others were labeled 0. The dataset was stratified and split into 80\% training and 20\% testing sets, maintaining label distribution across all splits.

\subsection{Classification and Evaluation}

The aggregated sample embeddings \( \mathbf{z}_S \in \mathbb{R}^{768} \) are passed to a downstream classifier to predict the target label (e.g., disease status, environmental type). We used two classifiers:

  \textbf{Fully Connected Neural Network (FCNN)}: A feedforward network~\cite{glorot2010understanding, glorot2011deep} with one 128-dimensional hidden layer, ReLU activation, and softmax output. The model was trained using cross-entropy loss for 10 epochs.

  \textbf{Random Forest (RF)}: A classical ensemble method~\cite{breiman2001random} with 100 estimators and class-balanced weighting.

Performance was measured using accuracy, macro-averaged precision, recall~\cite{salton1971smart}, and F1 score~\cite{vanrijsbergen1979information, sokolova2009systematic} on the held-out test set.

\section{Experiments}

We evaluate our model across three distinct microbiome classification tasks, leveraging datasets from the Qiita platform \cite{Gonzalez2018Qiita}, a centralized repository for standardized microbiome studies. Each task represents a unique biological or environmental prediction scenario, designed to test the utility of abundance-aware embedding in both clinical and ecological contexts.

Study identifiers refer to Qiita Study IDs, and if available, we cite associated original publications to ensure clarity and reproducibility.



\begin{table*}[h!]
\centering
\caption{FCNN and Random Forest classification results on soil vs. non-soil prediction.}
\label{tab:results_soil}
\begin{tabular}{lccccc}
\toprule
\textbf{Embedding Method} & \textbf{Classifier} & \textbf{Accuracy} & \textbf{Macro Precision} & \textbf{Macro Recall} & \textbf{Macro F1} \\
\midrule
Set Transformer           &FCNN & 0.4118 & 0.2059 & 0.5000 & 0.2917 \\
Average                   &FCNN & 0.4118 & 0.2059 & 0.5000 & 0.2917 \\
Weighted Average          &FCNN & 0.4118 & 0.2059 & 0.5000 & 0.2917 \\
Weighted Set Transformer  &FCNN & \textbf{0.5882} & \textbf{0.2941} & \textbf{0.5000} & \textbf{0.3704} \\
\midrule
Set Transformer           &Random Forest & 0.4118 & 0.2059 & 0.5000 & 0.2917 \\
Average                   &Random Forest & 0.4118 & 0.2059 & 0.5000 & 0.2917 \\
Weighted Average          &Random Forest & 0.4118 & 0.2059 & 0.5000 & 0.2917 \\
Weighted Set Transformer  &Random Forest & 0.4118 & 0.2059 & 0.5000 & 0.2917 \\
\bottomrule
\end{tabular}
\end{table*}

\subsection{Task 1: Bladder Microbiota Classification (Qiita Study 14245)}
This task investigates whether the urinary microbiome differs between tumor and non-tumor mucosa in bladder cancer patients \cite{qiita14245}. Prior research has suggested that certain microbial taxa, such as Actinobacteria, are more enriched in healthy bladder tissues and may play protective roles. In this study, we use paired tissue samples (tumor vs. adjacent non-tumor) to classify disease status based on microbial composition. Abundance plays a crucial role here, as specific low-abundance taxa (e.g., Enterococcus, Barnesiella) have been associated with tumor grade and subtypes. This task allows us to test whether embedding strategies can preserve and utilize these biologically meaningful abundance signals for predictive modeling.
Notably, this classification task operates on a relatively small dataset with only 116 samples but nearly 4,000 microbial features as shown in Table~\ref{tab:dataset_summary}, which exacerbates the risk of overfitting and highlights the need for embedding strategies that can compress high-dimensional inputs while preserving biologically salient information.

\subsection{Task 2: Acanthamoeba–Leptospira Co-Occurrence Prediction (Qiita Study 10442)}
This environmental task aims to predict whether free-living amoeba (FLA), such as Acanthamoeba, are associated with the presence of Leptospira—a pathogenic bacterium linked to waterborne disease outbreaks \cite{matthias2025fla_leptospira}. Understanding their co-occurrence is vital for assessing environmental reservoirs of infection. Samples were collected from freshwater environments in the tropics, and co-occurrence patterns may be subtle or rare, making abundance-aware modeling particularly useful in capturing weak signals that traditional averaging would miss. 
With 730 samples but over 300,000 microbial features as shown in Table~\ref{tab:dataset_summary}, this task presents an extreme high-dimensional setting where conventional classifiers struggle to generalize. The disproportion between feature space and label availability (656 positive vs. 74 negative) further compounds the difficulty, demanding approaches that can leverage abundance cues to filter signal from noise.

\subsection{Task 3: Soil vs. Non-Soil Environment Prediction (Qiita Studies 15573 and 1728)}
This task addresses a cross-study generalization challenge: determining whether a microbiome sample originates from a soil-associated environment. We used surface swab and plankton samples from the Caribbean marine ecosystem study (Study 15573) \cite{qiita15573} as training data and tested on a separate urban soil-related dataset from an asphalt site (Study 1728) \cite{qiita1728}. The samples in Study 15573 reflect natural marine surfaces rich in microbial diversity and possible pathogen contamination (e.g., Philaster clade associated with sea urchin die-offs), while Study 1728 includes engineered environments with distinct microbial compositions. This task simulates domain shift and tests whether abundance-aware embeddings are more robust to environmental heterogeneity.
This domain adaptation task is particularly challenging due to the extremely limited number of training samples (only 27 in Study 15573) and the vast microbial feature space as shown in Table~\ref{tab:dataset_summary}, making it a low-resource, high-dimensional learning scenario. The test set (Study 1728) also remains small (17 samples), further stressing the need for generalizable and efficient representation learning.

\section{Dataset Details}

We provide detailed metadata for the four datasets used across our three classification tasks. Table~\ref{tab:dataset_summary} summarizes the number of samples, label distribution, and environment types following the EMPO 3 ontology~\cite{thompson2017communal}.

\paragraph{Study 14245 – Bladder Microbiota.}
This clinical dataset includes 116 human samples from bladder mucosa: 52 labeled as "bladder mucosa" and 64 as "tumor primary - bladder." Labels were binarized into nontumor (0) and tumor (1). Each sample contains thousands of non-zero abundance nucleotide sequences derived from paired tissue biopsies. The dataset exhibits balanced class labels and captures microbial richness differences between tumor and non-tumor environments, as noted in prior studies.

\paragraph{Study 10442 – Leptospira Co-Occurrence.}
This environmental dataset contains 788 freshwater samples tested for the co-occurrence of pathogenic Leptospira with free-living amoebae. We discarded 58 "not applicable" samples and binarized the remaining 730 samples: 656 labeled as co-occurring (1) and 74 as not (0). The labels were derived from quantitative association scores (0.36 and 0.28) rounded into binary classes. Samples come from diverse surface water habitats with varying microbial compositions.

\paragraph{Study 15573 – Marine Surface Samples.}
This dataset consists of 27 training samples collected from marine surfaces including coral, algae, and sponges. The EMPO 3 annotations include Plant (saline) (6), Animal (saline) (17), Aqueous (saline) (1), and Solid (non-saline) (3). For soil classification purposes, we labeled only Solid (non-saline) samples as soil (label 1), and all other types as non-soil (label 0), resulting in 3 positive and 24 negative samples.

\paragraph{Study 1728 – Urban Asphalt Samples.}
This test set for our cross-domain soil prediction task includes 17 samples collected from asphalt surfaces and surrounding non-saline nearby water bodies. According to EMPO 3, it includes Solid (non-saline) (10) and Aqueous (non-saline) (7). As in Study 15573, we treated Solid (non-saline) samples as soil-associated (label 1), and Aqueous (non-saline) as non-soil (label 0), resulting in 10 soil and 7 non-soil samples. The environmental contrast between this and Study 15573 poses a challenging domain shift scenario.

\begin{table*}[h!]
\centering
\caption{Summary of datasets used across the three tasks. Labels were determined using either study-specific metadata or EMPO 3 annotations when available.}
\resizebox{\linewidth}{!}{%
\begin{tabular}{l l c  c c l c c}
\toprule
Study ID & Description & \# Samples & \# Features & Label 1 / 0 & Sample Type (EMPO 3 or Study Metadata) & Use & Domain \\
\midrule
14245 & Bladder mucosa (tumor vs. non-tumor) & 116 &3966 & 64 / 52 & 52 bladder mucosa, 64 tumor tissue (study metadata) & All & Clinical \\
10442 & Leptospira–Amoeba Co-occurrence & 730  &300176 & 656 / 74 & Co-occurrence: 0.36 (1), 0.28 (0), 58 N/A excluded (study metadata) & All & Environmental \\
15573 & Marine surfaces (coral/algae) & 27 &30601 & 3 / 24 & 6 Plant, 17 Animal, 1 Aqueous (saline), 3 Solid (non-saline) (EMPO 3) & Train & Marine \\
1728 & Urban asphalt/water & 17 &9644  & 10 / 7 & 10 Solid (non-saline), 7 Aqueous (non-saline) (EMPO 3) & Test & Urban \\
\bottomrule
\end{tabular}%
}
\label{tab:dataset_summary}
\end{table*}

\section{Results}

We evaluate the performance of four embedding aggregation strategies including Average, Weighted Average, Set Transformer, and our proposed Abundance-Aware Set Transformer. Across three distinct microbiome classification tasks, we use both a fully connected neural network (FCNN) and a Random Forest (RF) classifier.

\subsection{Study 14245 (Bladder Microbiota).}
As shown in Table~\ref{tab:results_14245}, all embedding methods performed equally under the FCNN classifier, suggesting limited model capacity or insufficient signal under this architecture. However, Random Forest results revealed a modest improvement when abundance information was considered. The Abundance-Aware Set Transformer outperformed all other methods, achieving the highest accuracy (0.5833) and macro F1 score (0.5804).

\subsection{Study 10442 (Acanthamoeba–Leptospira Co-Occurrence).}
Table~\ref{tab:results_10442} demonstrates that the Abundance-Aware Set Transformer achieved perfect performance (accuracy and macro F1 of 1.0) under both classifiers. This reflects its ability to capture biologically meaningful co-occurrence patterns that are likely subtle or sparse in abundance. Other methods plateaued at around 92–96\% accuracy with significantly lower F1 scores.

\subsection{Studies 15573 and 1728 (Soil vs. Non-Soil Prediction).}
This task involved cross-study generalization across marine and terrestrial environments. As seen in Table~\ref{tab:results_soil}, the FCNN classifier benefited from abundance-aware embeddings, with the Abundance-Aware Set Transformer achieving 0.5882 accuracy and 0.3704 macro F1. In contrast, the Random Forest classifier yielded uniformly low scores across all methods, likely due to domain shift and the small number of samples.

These results suggest that for this challenging cross-study task with substantial environmental and sampling differences between the training (Study~15573~\cite{qiita15573}) and test (Study~1728~\cite{qiita1728}) datasets, abundance-aware embedding offers some benefit under deep learning architectures, but traditional classifiers like Random Forest struggle to capture discriminative information. The predictive performance was inherently limited by the very small number of training and test samples. Larger and more diverse datasets would likely yield better results. Nevertheless, even under these constraints, the Weighted Set Transformer consistently outperformed the other embedding strategies.

\subsection{Summary}
Across all tasks, the \textit{Abundance-Aware Set Transformer} consistently outperformed baseline aggregation strategies, especially under deep neural architectures. These results highlight the importance of incorporating abundance information when constructing microbiome sample embeddings and suggest that attention-based methods can effectively leverage such quantitative metadata.

\section{Embedding Visualization and Interpretation}

In this section, we qualitatively analyze the sample-level embeddings produced by each aggregation strategy through dimensionality reduction techniques~\cite{wold1987principal, borg2005modern, kohonen2001self}. Due to their ability to capture non-linear structure and local neighborhood relationships, we focus on comparing t-SNE~\cite{van2008visualizing} and UMAP~\cite{mcinnes2018umap} to evaluate whether the learned representations separate biologically meaningful classes and preserve local/global structure. These visualizations provide interpretability insights that support the quantitative results reported in Section 6.

\begin{figure*}[h!]
    \centering    
    \includegraphics[width=1\linewidth]{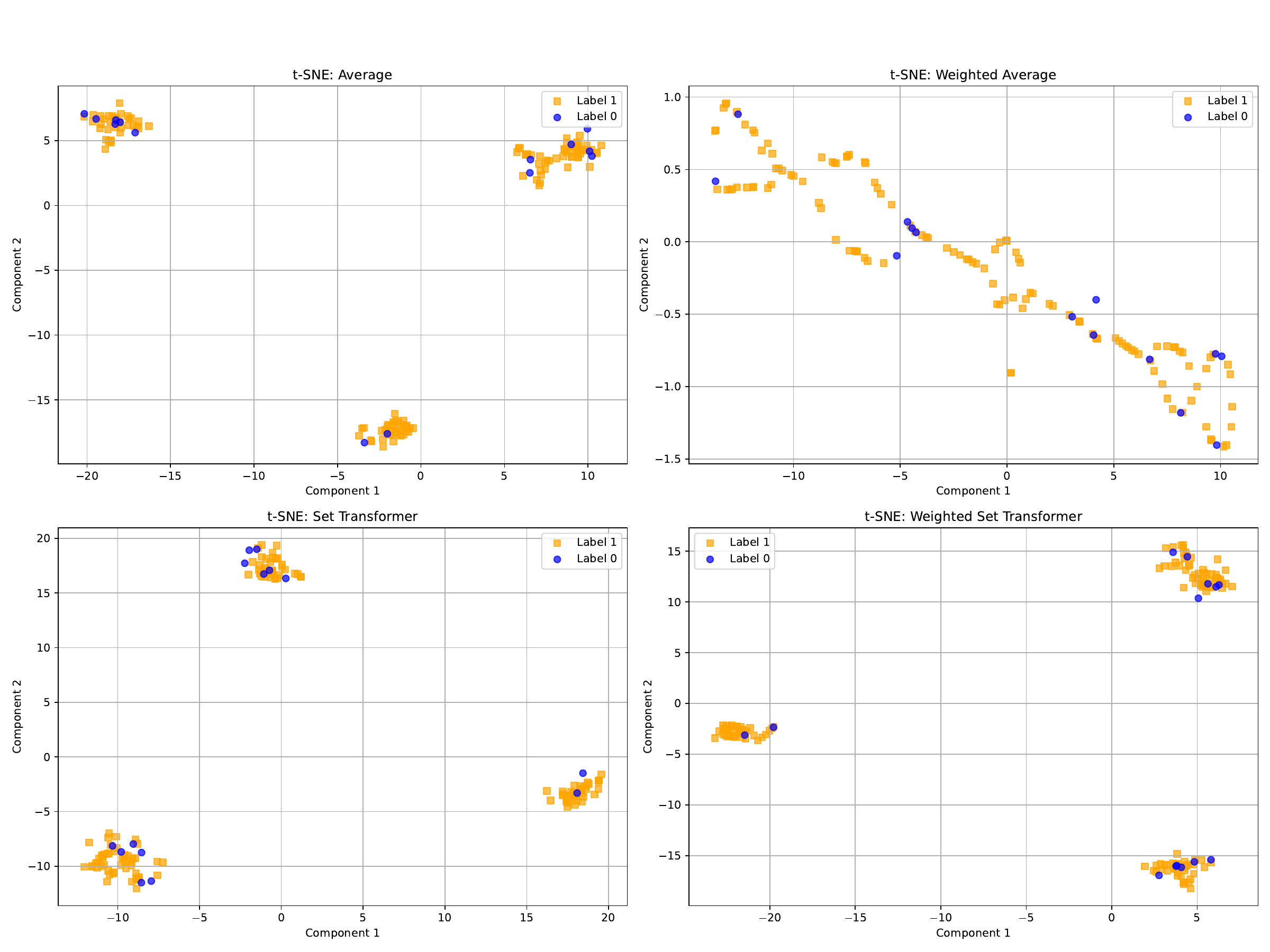}    
    \caption{t-SNE visualization of training set embeddings under different aggregation strategies. Blue circles denote Label 0 and orange squares denote Label 1. The Set Transformer and Weighted Set Transformer achieve better separation between labels compared to average-based methods.}
    \label{fig:tsne-visualization}
\end{figure*}

\subsection{t-SNE Visualization}

We applied t-distributed stochastic neighbor embedding (t-SNE)~\cite{van2008visualizing} to project the 768-dimensional sample embeddings into a two-dimensional space for visualization. While t-SNE is effective at preserving local neighborhood structures, it may distort global geometry and is sensitive to hyperparameters such as perplexity and learning rate.

As shown in Figure~\ref{fig:tsne-visualization}, embeddings produced by average-based aggregation methods (top row) exhibit significant overlap between Label 0 and Label 1, indicating poor inter-class separation. In contrast, attention-based methods such as the Set Transformer and Weighted Set Transformer (bottom row) achieve more distinct clusters, with clearer boundaries between the two labels. This suggests that transformer-based aggregation strategies can better capture class-relevant structures in the embedding space, even when visualized using t-SNE.

This visualization was conducted using training data from \textit{Study 10442}~\cite{matthias2025fla_leptospira}, which involves predicting co-occurrence between Acanthamoeba and Leptospira. This task is biologically important, as the abundance patterns of co-occurring species are often sparse and subtle, making it a strong testbed for evaluating the representational capacity of embedding strategies.

\begin{figure*}
    \centering    
    \includegraphics[width=1\linewidth]{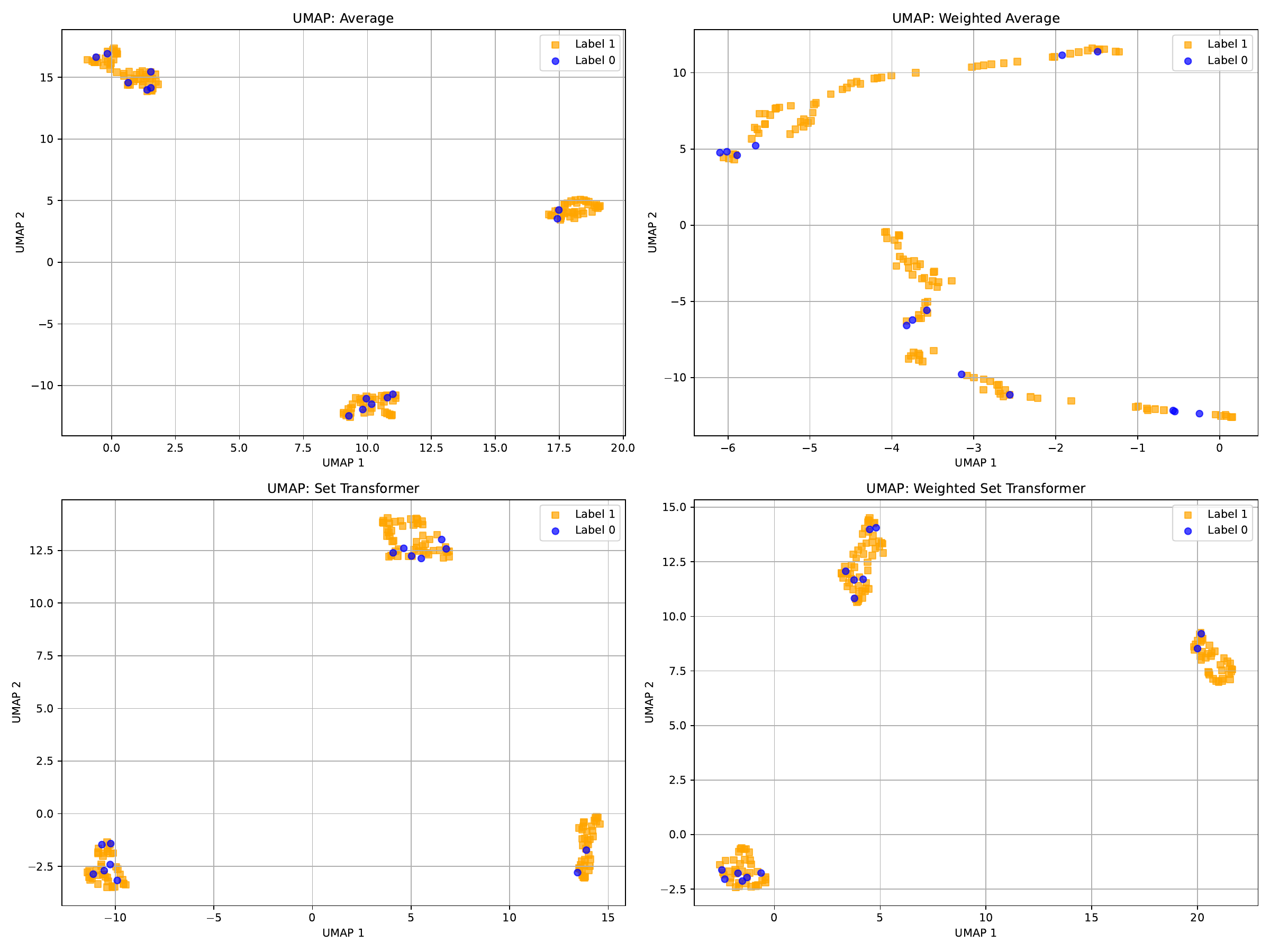}
    \caption{UMAP visualization of training set embeddings under different aggregation strategies. Blue circles denote Label 0 and orange squares denote Label 1. The Set Transformer and Weighted Set Transformer achieve better separation between labels compared to average-based methods.}
    \label{fig:umap-visualization}
\end{figure*}

\subsection{UMAP Visualization}

To qualitatively evaluate the behavior of each aggregation strategy, we visualized the training set embeddings using Uniform Manifold Approximation and Projection (UMAP)~\cite{mcinnes2018umap}. Figure~\ref{fig:umap-visualization} compares four embedding methods—Average pooling, Weighted Average pooling, Set Transformer, and Weighted Set Transformer—applied to the same dataset. Each point represents a microbiome sample, colored by its ground-truth label.

Among the four methods, the Set Transformer and the Weighted Set Transformer produced clearer separation between the two label groups. In particular, the Set Transformer yielded two distinct clusters with minimal overlap, suggesting that self-attention can effectively capture label-relevant structure from sets of sequence embeddings. Similarly, the Weighted Set Transformer showed compact groupings of same-label samples and less inter-class mixing, while additionally incorporating abundance information in a soft and biologically meaningful way. 
This visualization was also conducted using training data from \textit{Study 10442}~\cite{matthias2025fla_leptospira}.

In contrast, both Average pooling and Weighted Average pooling resulted in more entangled representations. Average pooling produced three loose clusters where class labels were often intermixed, limiting interpretability. Weighted Average, while incorporating abundance information, formed a stretched, nonlinear structure, which appeared to reflect some latent ordering but did not separate the classes as cleanly as attention-based approaches.

We also found UMAP to be more suitable than t-SNE for this visualization, as it preserved global structures and provided more stable inter-cluster relationships. While t-SNE emphasizes local neighborhood preservation, UMAP's manifold-preserving nature makes it preferable for interpreting embedding distributions at the sample level, particularly in biological data where both local and global context are relevant.

Taken together, these visualizations support the quantitative findings presented in previous sections. Attention-based aggregation methods, especially when abundance information is incorporated, not only yield higher classification performance but also produce embeddings that are more interpretable and discriminative in low-dimensional projections.

\section{Discussion}

Our experiments demonstrate that incorporating taxonomic abundance into sample level embedding strategies yields substantial improvements in both classification performance and interpretability. Traditional aggregation methods such as average pooling or unweighted Set Transformers fail to capture subtle but biologically meaningful signals, particularly in tasks where low-abundance taxa play a discriminative role. For example, in the Leptospira co-occurrence prediction task (Study 10442), only the Abundance Aware Set Transformer achieved perfect classification, suggesting its superior capacity to encode fine-grained compositional information.

Incorporating abundance via soft-weighted attention or sequence repetition allows attention-based models to emphasize taxa with higher biological relevance. Notably, the weighted Set Transformer outperformed all other aggregation strategies in the challenging cross-study soil classification task (Studies 15573 and 1728), despite the small number of training examples and domain shift. This highlights the method’s robustness and generalizability in scenarios with high heterogeneity and limited data.

Interestingly, we observed that fully connected neural networks benefited more from abundance-aware embeddings than traditional classifiers such as Random Forests, especially in low-resource or domain-shifted contexts. This suggests that deep learning architectures may be better suited to exploit the nuanced representations produced by Transformer based aggregation.

Our visualization analysis further supports these findings. Both t-SNE and UMAP projections reveal that set transformer based embeddings form more coherent and separable clusters, underscoring their ability to capture latent structure in the microbiome sample space. Importantly, abundance aware methods produce more compact clusters with reduced inter class overlap, providing better interpretability for downstream biological insights.

However, our approach still has limitations. First, the repetition-based encoding strategy, while effective, may introduce computational overhead in extremely large datasets. Second, our experiments were limited to three classification tasks; additional studies across more diverse environmental and clinical settings are needed to fully validate generalizability. Finally, while we focused on abundance at the sequence level, future work could explore multi-level abundance encoding, including taxonomic or functional hierarchies.

\section{Conclusion}

We present an abundance aware Set Transformer for microbiome sample embedding that integrates quantitative taxonomic information into attention based aggregation. By leveraging either soft attention weighting or repetition based encoding, our method captures the biological importance of sequence abundance without modifying the core Transformer architecture. Experiments across three diverse microbiome classification tasks show consistent performance gains, including perfect classification in subtle co-occurrence prediction and improved generalization in cross study scenarios.

In addition to quantitative gains, our embedding strategy has interpretability, as shown through low dimensional visualizations. These results suggest that attention based models, when informed by biological abundance, provide a useful and scalable framework for microbiome representation learning.

Future work will explore extending this framework to multi-omics integration, hierarchical abundance encoding, and applications in longitudinal microbiome analysis. Our approach opens new possibilities for biologically informed deep learning in microbiome science and beyond.



\bibliographystyle{ACM-Reference-Format}
\bibliography{sample-base}


\begin{thebibliography}{39}


\ifx \showCODEN    \undefined \def \showCODEN     #1{\unskip}     \fi
\ifx \showISBNx    \undefined \def \showISBNx     #1{\unskip}     \fi
\ifx \showISBNxiii \undefined \def \showISBNxiii  #1{\unskip}     \fi
\ifx \showISSN     \undefined \def \showISSN      #1{\unskip}     \fi
\ifx \showLCCN     \undefined \def \showLCCN      #1{\unskip}     \fi
\ifx \shownote     \undefined \def \shownote      #1{#1}          \fi
\ifx \showarticletitle \undefined \def \showarticletitle #1{#1}   \fi
\ifx \showURL      \undefined \def \showURL       {\relax}        \fi
\providecommand\bibfield[2]{#2}
\providecommand\bibinfo[2]{#2}
\providecommand\natexlab[1]{#1}
\providecommand\showeprint[2][]{arXiv:#2}

\bibitem[Ackerman and Baum(2022)]%
        {qiita1728}
\bibfield{author}{\bibinfo{person}{Gail Ackerman} {and} \bibinfo{person}{Marc Baum}.} \bibinfo{year}{2022}\natexlab{}.
\newblock \bibinfo{title}{Baum asphalt 1st submission (Baum asphalt study)}.
\newblock \bibinfo{howpublished}{\url{https://qiita.ucsd.edu/study/description/1728}}.
\newblock
\newblock
\shownote{Qiita Study ID: 1728}.


\bibitem[Arora et~al\mbox{.}(2017)]%
        {arora2017simple}
\bibfield{author}{\bibinfo{person}{Sanjeev Arora}, \bibinfo{person}{Yingyu Liang}, {and} \bibinfo{person}{Tengyu Ma}.} \bibinfo{year}{2017}\natexlab{}.
\newblock \showarticletitle{A Simple but Tough-to-Beat Baseline for Sentence Embeddings}. In \bibinfo{booktitle}{\emph{Proceedings of the 5th International Conference on Learning Representations (ICLR)}}.
\newblock
\urldef\tempurl%
\url{https://openreview.net/forum?id=SyK00v5xx}
\showURL{%
\tempurl}


\bibitem[Arumugam et~al\mbox{.}(2011)]%
        {arumugam2011enterotypes}
\bibfield{author}{\bibinfo{person}{Manimozhiyan Arumugam}, \bibinfo{person}{Jeroen Raes}, \bibinfo{person}{Eric Pelletier}, \bibinfo{person}{Denis Le~Paslier}, \bibinfo{person}{Takuji Yamada}, \bibinfo{person}{Daniel~R. Mende}, \bibinfo{person}{Gabriel~R. Fernandes}, \bibinfo{person}{Julien Tap}, \bibinfo{person}{Thomas Bruls}, \bibinfo{person}{Jean-Michel Batto}, \bibinfo{person}{Marcelo Bertalan}, \bibinfo{person}{Natalia Borruel}, \bibinfo{person}{Francesc Casellas}, \bibinfo{person}{Leyden Fernandez}, \bibinfo{person}{Laurent Gautier}, \bibinfo{person}{Torben Hansen}, \bibinfo{person}{Masahira Hattori}, \bibinfo{person}{Tetsuya Hayashi}, \bibinfo{person}{Michiel Kleerebezem}, \bibinfo{person}{Ken Kurokawa}, \bibinfo{person}{Marion Leclerc}, \bibinfo{person}{Florence Levenez}, \bibinfo{person}{Chaysavanh Manichanh}, \bibinfo{person}{H.~Bjørn Nielsen}, \bibinfo{person}{Trine Nielsen}, \bibinfo{person}{Nicolas Pons}, \bibinfo{person}{Julie Poulain}, \bibinfo{person}{Junjie Qin}, \bibinfo{person}{Thomas
  Sicheritz-Pontén}, \bibinfo{person}{Sebastian Tims}, \bibinfo{person}{David Torrents}, \bibinfo{person}{Edgardo Ugarte}, \bibinfo{person}{Erwin~G. Zoetendal}, \bibinfo{person}{Jun Wang}, \bibinfo{person}{Francisco Guarner}, \bibinfo{person}{Oluf Pedersen}, \bibinfo{person}{Willem~M. de Vos}, \bibinfo{person}{Søren Brunak}, \bibinfo{person}{Joel Doré}, \bibinfo{person}{MetaHIT Consortium}, \bibinfo{person}{Jean Weissenbach}, \bibinfo{person}{S.~Dusko Ehrlich}, {and} \bibinfo{person}{Peer Bork}.} \bibinfo{year}{2011}\natexlab{}.
\newblock \showarticletitle{Enterotypes of the human gut microbiome}.
\newblock \bibinfo{journal}{\emph{Nature}}  \bibinfo{volume}{473} (\bibinfo{year}{2011}), \bibinfo{pages}{174--180}.
\newblock
\href{https://doi.org/10.1038/nature09944}{doi:\nolinkurl{10.1038/nature09944}}


\bibitem[Bolyen et~al\mbox{.}(2019)]%
        {bolyen2019qiime2}
\bibfield{author}{\bibinfo{person}{Evan Bolyen}, \bibinfo{person}{Jai~Ram Rideout}, \bibinfo{person}{Matthew~R. Dillon}, \bibinfo{person}{Nicholas~A. Bokulich}, \bibinfo{person}{Christian~C. Abnet}, \bibinfo{person}{Gabriel~A. Al-Ghalith}, \bibinfo{person}{Harriet Alexander}, \bibinfo{person}{Eric~J. Alm}, \bibinfo{person}{Manimozhiyan Arumugam}, \bibinfo{person}{Francesco Asnicar}, \bibinfo{person}{Yang Bai}, \bibinfo{person}{Jordan~E. Bisanz}, \bibinfo{person}{Kyle Bittinger}, \bibinfo{person}{Asker Brejnrod}, \bibinfo{person}{Colin~J. Brislawn}, \bibinfo{person}{C.~Titus Brown}, \bibinfo{person}{Benjamin~J. Callahan}, \bibinfo{person}{Andrés~Mauricio Caraballo-Rodríguez}, \bibinfo{person}{John Chase}, \bibinfo{person}{Emily~K. Cope}, \bibinfo{person}{Ricardo Da~Silva}, \bibinfo{person}{Christian Diener}, \bibinfo{person}{Pieter~C. Dorrestein}, \bibinfo{person}{Gavin~M. Douglas}, \bibinfo{person}{Daniel~M. Durall}, \bibinfo{person}{Claire Duvallet}, \bibinfo{person}{Christian~F. Edwardson},
  \bibinfo{person}{Madeleine Ernst}, \bibinfo{person}{Mehrbod Estaki}, \bibinfo{person}{Jennifer Fouquier}, \bibinfo{person}{Julia~M. Gauglitz}, \bibinfo{person}{Sean~M. Gibbons}, \bibinfo{person}{Deanna~L. Gibson}, \bibinfo{person}{Antonio Gonzalez}, \bibinfo{person}{Kestrel Gorlick}, \bibinfo{person}{Jiarong Guo}, \bibinfo{person}{Benjamin Hillmann}, \bibinfo{person}{Susan Holmes}, \bibinfo{person}{Hannes Holste}, \bibinfo{person}{Curtis Huttenhower}, \bibinfo{person}{Gavin~A. Huttley}, \bibinfo{person}{Stefan Janssen}, \bibinfo{person}{Alan~K. Jarmusch}, \bibinfo{person}{Lingjing Jiang}, \bibinfo{person}{Benjamin~D. Kaehler}, \bibinfo{person}{Kyo~Bin Kang}, \bibinfo{person}{Christopher~R. Keefe}, \bibinfo{person}{Paul Keim}, \bibinfo{person}{Scott~T. Kelley}, \bibinfo{person}{Dan Knights}, \bibinfo{person}{Irina Koester}, \bibinfo{person}{Tomasz Kosciolek}, \bibinfo{person}{Jorden Kreps}, \bibinfo{person}{Morgan G.~I. Langille}, \bibinfo{person}{Joslynn Lee}, \bibinfo{person}{Ruth Ley},
  \bibinfo{person}{Yong-Xin Liu}, \bibinfo{person}{Erikka Loftfield}, \bibinfo{person}{Catherine Lozupone}, \bibinfo{person}{Massoud Maher}, \bibinfo{person}{Clarisse Marotz}, \bibinfo{person}{Bryan~D. Martin}, \bibinfo{person}{Daniel McDonald}, \bibinfo{person}{Lauren~J. McIver}, \bibinfo{person}{Alexey~V. Melnik}, \bibinfo{person}{Jessica~L. Metcalf}, \bibinfo{person}{Sydney~C. Morgan}, \bibinfo{person}{Jamie~T. Morton}, \bibinfo{person}{Ahmad~Turan Naimey}, \bibinfo{person}{Jose~A. Navas-Molina}, \bibinfo{person}{Louis~Felix Nothias}, \bibinfo{person}{Stephanie~B. Orchanian}, \bibinfo{person}{Talima Pearson}, \bibinfo{person}{Samuel~L. Peoples}, \bibinfo{person}{Daniel Petras}, \bibinfo{person}{Mary~Lai Preuss}, \bibinfo{person}{Elmar Pruesse}, \bibinfo{person}{Lasse~Buur Rasmussen}, \bibinfo{person}{Adam Rivers}, \bibinfo{person}{Michael~S. Robeson~II}, \bibinfo{person}{Patrick Rosenthal}, \bibinfo{person}{Nicola Segata}, \bibinfo{person}{Michael Shaffer}, \bibinfo{person}{Arron Shiffer},
  \bibinfo{person}{Rashmi Sinha}, \bibinfo{person}{Se~Jin Song}, \bibinfo{person}{John~R. Spear}, \bibinfo{person}{Austin~D. Swafford}, \bibinfo{person}{Luke~R. Thompson}, \bibinfo{person}{Pedro~J. Torres}, \bibinfo{person}{Pauline Trinh}, \bibinfo{person}{Anupriya Tripathi}, \bibinfo{person}{Peter~J. Turnbaugh}, \bibinfo{person}{Sabah Ul-Hasan}, \bibinfo{person}{Justin J.~J. van~der Hooft}, \bibinfo{person}{Fernando Vargas}, \bibinfo{person}{Yoshiki Vázquez-Baeza}, \bibinfo{person}{Emily Vogtmann}, \bibinfo{person}{Max von Hippel}, \bibinfo{person}{William Walters}, \bibinfo{person}{Yunhu Wan}, \bibinfo{person}{Mingxun Wang}, \bibinfo{person}{Jonathan Warren}, \bibinfo{person}{Kyle~C. Weber}, \bibinfo{person}{Charles H.~D. Williamson}, \bibinfo{person}{Amy~D. Willis}, \bibinfo{person}{Zhenjiang~Zech Xu}, \bibinfo{person}{Jesse~R. Zaneveld}, \bibinfo{person}{Yilong Zhang}, \bibinfo{person}{Qiyun Zhu}, \bibinfo{person}{Rob Knight}, {and} \bibinfo{person}{J.~Gregory Caporaso}.} \bibinfo{year}{2019}\natexlab{}.
\newblock \showarticletitle{Reproducible, interactive, scalable and extensible microbiome data science using QIIME 2}.
\newblock \bibinfo{journal}{\emph{Nature Biotechnology}}  \bibinfo{volume}{37} (\bibinfo{year}{2019}), \bibinfo{pages}{852--857}.
\newblock
\href{https://doi.org/10.1038/s41587-019-0209-9}{doi:\nolinkurl{10.1038/s41587-019-0209-9}}
\newblock
\shownote{Correspondence}.


\bibitem[Borg and Groenen(2005)]%
        {borg2005modern}
\bibfield{author}{\bibinfo{person}{Ingwer Borg} {and} \bibinfo{person}{Patrick J.~F. Groenen}.} \bibinfo{year}{2005}\natexlab{}.
\newblock \bibinfo{booktitle}{\emph{Modern Multidimensional Scaling: Theory and Applications} (\bibinfo{edition}{2nd} ed.)}.
\newblock \bibinfo{publisher}{Springer}.
\newblock
\showISBNx{978-0-387-28980-5}
\href{https://doi.org/10.1007/0-387-28981-X}{doi:\nolinkurl{10.1007/0-387-28981-X}}


\bibitem[Breiman(2001)]%
        {breiman2001random}
\bibfield{author}{\bibinfo{person}{Leo Breiman}.} \bibinfo{year}{2001}\natexlab{}.
\newblock \showarticletitle{Random Forests}.
\newblock \bibinfo{journal}{\emph{Machine Learning}} \bibinfo{volume}{45}, \bibinfo{number}{1} (\bibinfo{year}{2001}), \bibinfo{pages}{5--32}.
\newblock
\href{https://doi.org/10.1023/A:1010933404324}{doi:\nolinkurl{10.1023/A:1010933404324}}


\bibitem[Dalla-Torre et~al\mbox{.}(2023)]%
        {dalla-torre2023nucleotide}
\bibfield{author}{\bibinfo{person}{Hugo Dalla-Torre}, \bibinfo{person}{Liam Gonzalez}, \bibinfo{person}{Javier Mendoza~Revilla}, \bibinfo{person}{Nicolas Lopez~Carranza}, \bibinfo{person}{Adam~Henryk Grzywaczewski}, \bibinfo{person}{Francesco Oteri}, \bibinfo{person}{Christian Dallago}, \bibinfo{person}{Evan Trop}, \bibinfo{person}{Hassan Sirelkhatim}, \bibinfo{person}{Guillaume Richard}, \bibinfo{person}{Marcin Skwark}, \bibinfo{person}{Karim Beguir}, \bibinfo{person}{Marie Lopez}, {and} \bibinfo{person}{Thomas Pierrot}.} \bibinfo{year}{2023}\natexlab{}.
\newblock \showarticletitle{The Nucleotide Transformer: Building and Evaluating Robust Foundation Models for Human Genomics}.
\newblock \bibinfo{journal}{\emph{bioRxiv}} (\bibinfo{year}{2023}).
\newblock
\href{https://doi.org/10.1101/2023.01.11.523679}{doi:\nolinkurl{10.1101/2023.01.11.523679}}


\bibitem[Glorot and Bengio(2010)]%
        {glorot2010understanding}
\bibfield{author}{\bibinfo{person}{Xavier Glorot} {and} \bibinfo{person}{Yoshua Bengio}.} \bibinfo{year}{2010}\natexlab{}.
\newblock \showarticletitle{Understanding the difficulty of training deep feedforward neural networks}. In \bibinfo{booktitle}{\emph{Proceedings of the 13th International Conference on Artificial Intelligence and Statistics (AISTATS)}} \emph{(\bibinfo{series}{Proceedings of Machine Learning Research (PMLR)}, Vol.~\bibinfo{volume}{9})}, \bibfield{editor}{\bibinfo{person}{Yee~Whye Teh} {and} \bibinfo{person}{Mike Titterington}} (Eds.). \bibinfo{publisher}{PMLR}, \bibinfo{address}{Chia Laguna Resort, Sardinia, Italy}, \bibinfo{pages}{249--256}.
\newblock
\urldef\tempurl%
\url{http://proceedings.mlr.press/v9/glorot10a.html}
\showURL{%
\tempurl}


\bibitem[Glorot et~al\mbox{.}(2011)]%
        {glorot2011deep}
\bibfield{author}{\bibinfo{person}{Xavier Glorot}, \bibinfo{person}{Antoine Bordes}, {and} \bibinfo{person}{Yoshua Bengio}.} \bibinfo{year}{2011}\natexlab{}.
\newblock \showarticletitle{Deep Sparse Rectifier Neural Networks}. In \bibinfo{booktitle}{\emph{Proceedings of the 14th International Conference on Artificial Intelligence and Statistics (AISTATS)}} \emph{(\bibinfo{series}{Proceedings of Machine Learning Research (PMLR)}, Vol.~\bibinfo{volume}{15})}, \bibfield{editor}{\bibinfo{person}{Geoffrey Gordon}, \bibinfo{person}{David Dunson}, {and} \bibinfo{person}{Miroslav Dudík}} (Eds.). \bibinfo{publisher}{PMLR}, \bibinfo{address}{Fort Lauderdale, FL, USA}, \bibinfo{pages}{315--323}.
\newblock
\urldef\tempurl%
\url{http://proceedings.mlr.press/v15/glorot11a.html}
\showURL{%
\tempurl}


\bibitem[Gonzalez et~al\mbox{.}(2018)]%
        {Gonzalez2018Qiita}
\bibfield{author}{\bibinfo{person}{Antonio Gonzalez}, \bibinfo{person}{Jose~A. Navas-Molina}, \bibinfo{person}{Tomasz Kosciolek}, \bibinfo{person}{Daniel McDonald}, \bibinfo{person}{Yoshiki Vázquez-Baeza}, \bibinfo{person}{Gail Ackermann}, \bibinfo{person}{Jeff DeReus}, \bibinfo{person}{Stefan Janssen}, \bibinfo{person}{Austin~D. Swafford}, \bibinfo{person}{Stephanie~B. Orchanian}, \bibinfo{person}{Jon~G. Sanders}, \bibinfo{person}{Joshua Shorenstein}, \bibinfo{person}{Hannes Holste}, \bibinfo{person}{Semar Petrus}, \bibinfo{person}{Adam Robbins-Pianka}, \bibinfo{person}{Colin~J. Brislawn}, \bibinfo{person}{Mingxun Wang}, \bibinfo{person}{Jai~Ram Rideout}, \bibinfo{person}{Evan Bolyen}, \bibinfo{person}{Matthew Dillon}, \bibinfo{person}{J.~Gregory Caporaso}, \bibinfo{person}{Pieter~C. Dorrestein}, {and} \bibinfo{person}{Rob Knight}.} \bibinfo{year}{2018}\natexlab{}.
\newblock \showarticletitle{Qiita: rapid, web-enabled microbiome meta-analysis}.
\newblock \bibinfo{journal}{\emph{Nature Methods}} \bibinfo{volume}{15}, \bibinfo{number}{10} (\bibinfo{year}{2018}), \bibinfo{pages}{796--798}.
\newblock
\href{https://doi.org/10.1038/s41592-018-0141-9}{doi:\nolinkurl{10.1038/s41592-018-0141-9}}


\bibitem[Hewson and vcbycedin@gmail.com(2024)]%
        {qiita15573}
\bibfield{author}{\bibinfo{person}{Ian Hewson} {and} \bibinfo{person}{vcbycedin@gmail.com}.} \bibinfo{year}{2024}\natexlab{}.
\newblock \bibinfo{title}{Detection of the Diadema antillarum scuticociliatosis Philaster clade on sympatric metazoa, plankton, and abiotic surfaces and assessment for its potential reemergence (DaScPc on Prevalence Marine Surfaces)}.
\newblock \bibinfo{howpublished}{\url{https://qiita.ucsd.edu/study/description/15573}}.
\newblock
\href{https://doi.org/10.3354/meps14763}{doi:\nolinkurl{10.3354/meps14763}}
\newblock
\shownote{Qiita Study ID: 15573}.


\bibitem[II et~al\mbox{.}(2025)]%
        {Ludwig2025}
\bibfield{author}{\bibinfo{person}{David W.~Ludwig II}, \bibinfo{person}{Christopher Guptil}, \bibinfo{person}{Nicholas~R. Alexander}, \bibinfo{person}{Kateryna Zhalnina}, \bibinfo{person}{Edi M.-L. Wipf}, \bibinfo{person}{Albina Khasanova}, \bibinfo{person}{Nicholas~A. Barber}, \bibinfo{person}{Wesley Swingley}, \bibinfo{person}{Donald~M. Walker}, {and} \bibinfo{person}{Joshua~L. Phillips}.} \bibinfo{year}{2025}\natexlab{}.
\newblock \showarticletitle{SetBERT: the deep learning platform for contextualized embeddings and explainable predictions from high-throughput sequencing}.
\newblock \bibinfo{journal}{\emph{Bioinformatics}} \bibinfo{volume}{41}, \bibinfo{number}{7} (\bibinfo{year}{2025}), \bibinfo{pages}{btaf370}.
\newblock
\href{https://doi.org/10.1093/bioinformatics/btaf370}{doi:\nolinkurl{10.1093/bioinformatics/btaf370}}


\bibitem[Ji et~al\mbox{.}(2021)]%
        {Ji2021DNABERT}
\bibfield{author}{\bibinfo{person}{Yanrong Ji}, \bibinfo{person}{Zhihan Zhou}, \bibinfo{person}{Han Liu}, {and} \bibinfo{person}{Ramana~V Davuluri}.} \bibinfo{year}{2021}\natexlab{}.
\newblock \showarticletitle{DNABERT: pre-trained Bidirectional Encoder Representations from Transformers model for DNA-language in genome}.
\newblock \bibinfo{journal}{\emph{Bioinformatics}} \bibinfo{volume}{37}, \bibinfo{number}{15} (\bibinfo{year}{2021}), \bibinfo{pages}{2112--2120}.
\newblock
\href{https://doi.org/10.1093/bioinformatics/btab083}{doi:\nolinkurl{10.1093/bioinformatics/btab083}}


\bibitem[Kohonen(2001)]%
        {kohonen2001self}
\bibfield{author}{\bibinfo{person}{Teuvo Kohonen}.} \bibinfo{year}{2001}\natexlab{}.
\newblock \bibinfo{booktitle}{\emph{Self-Organizing Maps} (\bibinfo{edition}{3rd} ed.)}. \bibinfo{series}{Springer Series in Information Sciences}, Vol.~\bibinfo{volume}{30}.
\newblock \bibinfo{publisher}{Springer}.
\newblock
\showISBNx{978-3-540-67921-9}
\href{https://doi.org/10.1007/978-3-642-56927-2}{doi:\nolinkurl{10.1007/978-3-642-56927-2}}


\bibitem[Lee et~al\mbox{.}(2019)]%
        {lee2019set}
\bibfield{author}{\bibinfo{person}{Juho Lee}, \bibinfo{person}{Yoonho Lee}, \bibinfo{person}{Jungtaek Kim}, \bibinfo{person}{Adam~R. Kosiorek}, \bibinfo{person}{Seungjin Choi}, {and} \bibinfo{person}{Yee~Whye Teh}.} \bibinfo{year}{2019}\natexlab{}.
\newblock \showarticletitle{Set Transformer: A Framework for Attention-based Permutation-Invariant Neural Networks}. In \bibinfo{booktitle}{\emph{Proceedings of the 36th International Conference on Machine Learning}} \emph{(\bibinfo{series}{Proceedings of Machine Learning Research}, Vol.~\bibinfo{volume}{97})}. \bibinfo{publisher}{PMLR}, \bibinfo{address}{Long Beach, California, USA}, \bibinfo{pages}{3744--3753}.
\newblock
\urldef\tempurl%
\url{http://proceedings.mlr.press/v97/lee19d.html}
\showURL{%
\tempurl}


\bibitem[Lo and Marculescu(2019)]%
        {lo2019metann}
\bibfield{author}{\bibinfo{person}{Chieh Lo} {and} \bibinfo{person}{Radu Marculescu}.} \bibinfo{year}{2019}\natexlab{}.
\newblock \showarticletitle{MetaNN: accurate classification of host phenotypes from metagenomic data using neural networks}.
\newblock \bibinfo{journal}{\emph{BMC Bioinformatics}} \bibinfo{volume}{20}, \bibinfo{number}{Suppl 12} (\bibinfo{year}{2019}), \bibinfo{pages}{314}.
\newblock
\href{https://doi.org/10.1186/s12859-019-2864-0}{doi:\nolinkurl{10.1186/s12859-019-2864-0}}


\bibitem[Mandal et~al\mbox{.}(2015)]%
        {mandal2015analysis}
\bibfield{author}{\bibinfo{person}{Siddhartha Mandal}, \bibinfo{person}{Will Van~Treuren}, \bibinfo{person}{Richard~A. White}, \bibinfo{person}{Merete Eggesbø}, \bibinfo{person}{Rob Knight}, {and} \bibinfo{person}{Shyamal~D. Peddada}.} \bibinfo{year}{2015}\natexlab{}.
\newblock \showarticletitle{Analysis of composition of microbiomes: a novel method for studying microbial composition}.
\newblock \bibinfo{journal}{\emph{Microbial Ecology in Health and Disease}}  \bibinfo{volume}{26} (\bibinfo{year}{2015}), \bibinfo{pages}{27663}.
\newblock
\href{https://doi.org/10.3402/mehd.v26.27663}{doi:\nolinkurl{10.3402/mehd.v26.27663}}


\bibitem[Martino et~al\mbox{.}(2019)]%
        {martino2019novel}
\bibfield{author}{\bibinfo{person}{Cameron Martino}, \bibinfo{person}{James~T. Morton}, \bibinfo{person}{Clarisse~A. Marotz}, \bibinfo{person}{Luke~R. Thompson}, \bibinfo{person}{Anupriya Tripathi}, \bibinfo{person}{Rob Knight}, {and} \bibinfo{person}{Karsten Zengler}.} \bibinfo{year}{2019}\natexlab{}.
\newblock \showarticletitle{A Novel Sparse Compositional Technique Reveals Microbial Perturbations}.
\newblock \bibinfo{journal}{\emph{mSystems}} \bibinfo{volume}{4}, \bibinfo{number}{1} (\bibinfo{year}{2019}), \bibinfo{pages}{e00016--19}.
\newblock
\href{https://doi.org/10.1128/mSystems.00016-19}{doi:\nolinkurl{10.1128/mSystems.00016-19}}


\bibitem[Matthias and Lubar(2025)]%
        {matthias2025fla_leptospira}
\bibfield{author}{\bibinfo{person}{Michael Matthias} {and} \bibinfo{person}{Aristea Lubar}.} \bibinfo{year}{2025}\natexlab{}.
\newblock \bibinfo{title}{Free-Living Amoeba Reservoirs of Pathogenic Leptospira}.
\newblock \bibinfo{howpublished}{\url{https://qiita.ucsd.edu/study/description/10442}}.
\newblock
\newblock
\shownote{Qiita Study ID: 10442}.


\bibitem[McInnes et~al\mbox{.}(2018)]%
        {mcinnes2018umap}
\bibfield{author}{\bibinfo{person}{Leland McInnes}, \bibinfo{person}{John Healy}, \bibinfo{person}{Nathaniel Saul}, {and} \bibinfo{person}{Lukas Großberger}.} \bibinfo{year}{2018}\natexlab{}.
\newblock \showarticletitle{UMAP: Uniform Manifold Approximation and Projection for Dimension Reduction}.
\newblock \bibinfo{journal}{\emph{Journal of Open Source Software}} \bibinfo{volume}{3}, \bibinfo{number}{29} (\bibinfo{year}{2018}), \bibinfo{pages}{861}.
\newblock
\href{https://doi.org/10.21105/joss.00861}{doi:\nolinkurl{10.21105/joss.00861}}


\bibitem[Mendler et~al\mbox{.}(2024)]%
        {mendler2024microbiome}
\bibfield{author}{\bibinfo{person}{Isabella-Hilda Mendler}, \bibinfo{person}{Barbara Drossel}, {and} \bibinfo{person}{Marc-Thorsten H{\"u}tt}.} \bibinfo{year}{2024}\natexlab{}.
\newblock \showarticletitle{Microbiome abundance patterns as attractors and the implications for the inference of microbial interaction networks}.
\newblock \bibinfo{journal}{\emph{Physica A: Statistical Mechanics and its Applications}}  \bibinfo{volume}{639} (\bibinfo{year}{2024}).
\newblock
\href{https://doi.org/10.1016/j.physa.2024.129567}{doi:\nolinkurl{10.1016/j.physa.2024.129567}}


\bibitem[Mendoza-Revilla et~al\mbox{.}(2024)]%
        {mendoza2024foundational}
\bibfield{author}{\bibinfo{person}{Javier Mendoza-Revilla}, \bibinfo{person}{Evan Trop}, \bibinfo{person}{Liam Gonzalez}, \bibinfo{person}{Maša Roller}, \bibinfo{person}{Hugo Dalla-Torre}, \bibinfo{person}{Bernardo~P. de Almeida}, \bibinfo{person}{Guillaume Richard}, \bibinfo{person}{Jonathan Caton}, \bibinfo{person}{Nicolas~Lopez Carranza}, \bibinfo{person}{Marcin Skwark}, \bibinfo{person}{Alex Laterre}, \bibinfo{person}{Karim Beguir}, \bibinfo{person}{Thomas Pierrot}, {and} \bibinfo{person}{Marie Lopez}.} \bibinfo{year}{2024}\natexlab{}.
\newblock \showarticletitle{A foundational large language model for edible plant genomes}.
\newblock \bibinfo{journal}{\emph{Communications Biology}}  \bibinfo{volume}{7} (\bibinfo{year}{2024}), \bibinfo{pages}{835}.
\newblock
\href{https://doi.org/10.1038/s42003-024-06186-w}{doi:\nolinkurl{10.1038/s42003-024-06186-w}}


\bibitem[Mikolov et~al\mbox{.}(2013)]%
        {mikolov2013efficient}
\bibfield{author}{\bibinfo{person}{Tomas Mikolov}, \bibinfo{person}{Kai Chen}, \bibinfo{person}{Greg Corrado}, {and} \bibinfo{person}{Jeffrey Dean}.} \bibinfo{year}{2013}\natexlab{}.
\newblock \showarticletitle{Efficient Estimation of Word Representations in Vector Space}. In \bibinfo{booktitle}{\emph{Proceedings of the International Conference on Learning Representations (ICLR)}}.
\newblock
\urldef\tempurl%
\url{https://arxiv.org/abs/1301.3781}
\showURL{%
\tempurl}
\newblock
\shownote{Workshop Track}.


\bibitem[Monshizadeh et~al\mbox{.}(2025)]%
        {monshizadeh2025multitask}
\bibfield{author}{\bibinfo{person}{Mahsa Monshizadeh}, \bibinfo{person}{Yuhui Hong}, {and} \bibinfo{person}{Yuzhen Ye}.} \bibinfo{year}{2025}\natexlab{}.
\newblock \showarticletitle{Multitask knowledge-primed neural network for predicting missing metadata and host phenotype based on human microbiome}.
\newblock \bibinfo{journal}{\emph{Bioinformatics Advances}} \bibinfo{volume}{5}, \bibinfo{number}{1} (\bibinfo{year}{2025}), \bibinfo{pages}{vbae203}.
\newblock
\href{https://doi.org/10.1093/bioadv/vbae203}{doi:\nolinkurl{10.1093/bioadv/vbae203}}


\bibitem[Oh and Zhang(2020)]%
        {oh2020deepmicro}
\bibfield{author}{\bibinfo{person}{Min Oh} {and} \bibinfo{person}{Liqing Zhang}.} \bibinfo{year}{2020}\natexlab{}.
\newblock \showarticletitle{DeepMicro: deep representation learning for disease prediction based on microbiome data}.
\newblock \bibinfo{journal}{\emph{Scientific Reports}}  \bibinfo{volume}{10} (\bibinfo{year}{2020}), \bibinfo{pages}{6026}.
\newblock
\href{https://doi.org/10.1038/s41598-020-62929-2}{doi:\nolinkurl{10.1038/s41598-020-62929-2}}


\bibitem[Pasolli et~al\mbox{.}(2016)]%
        {pasolli2016mlmeta}
\bibfield{author}{\bibinfo{person}{Edoardo Pasolli}, \bibinfo{person}{Duy~Tin Truong}, \bibinfo{person}{Faizan Malik}, \bibinfo{person}{Levi Waldron}, {and} \bibinfo{person}{Nicola Segata}.} \bibinfo{year}{2016}\natexlab{}.
\newblock \showarticletitle{Machine Learning Meta-analysis of Large Metagenomic Datasets: Tools and Biological Insights}.
\newblock \bibinfo{journal}{\emph{PLOS Computational Biology}} \bibinfo{volume}{12}, \bibinfo{number}{7} (\bibinfo{year}{2016}), \bibinfo{pages}{e1004977}.
\newblock
\href{https://doi.org/10.1371/journal.pcbi.1004977}{doi:\nolinkurl{10.1371/journal.pcbi.1004977}}


\bibitem[{Qiita Study 14245}(2021)]%
        {qiita14245}
\bibfield{author}{\bibinfo{person}{{Qiita Study 14245}}.} \bibinfo{year}{2021}\natexlab{}.
\newblock \bibinfo{title}{Bladder microbiota in paired samples of tumor and nontumor mucosa}.
\newblock \bibinfo{howpublished}{\url{https://qiita.ucsd.edu/study/description/14245}}.
\newblock


\bibitem[Salton(1971)]%
        {salton1971smart}
\bibfield{author}{\bibinfo{person}{Gerard Salton}.} \bibinfo{year}{1971}\natexlab{}.
\newblock \bibinfo{booktitle}{\emph{The SMART Retrieval System: Experiments in Automatic Document Processing}}.
\newblock \bibinfo{publisher}{Prentice-Hall}, \bibinfo{address}{Upper Saddle River, NJ, USA}.
\newblock


\bibitem[Sankaran and Holmes(2019)]%
        {sankaran2019latent}
\bibfield{author}{\bibinfo{person}{Kris Sankaran} {and} \bibinfo{person}{Susan~P. Holmes}.} \bibinfo{year}{2019}\natexlab{}.
\newblock \showarticletitle{Latent variable modeling for the microbiome}.
\newblock \bibinfo{journal}{\emph{Biostatistics}} \bibinfo{volume}{20}, \bibinfo{number}{4} (\bibinfo{year}{2019}), \bibinfo{pages}{599--614}.
\newblock
\href{https://doi.org/10.1093/biostatistics/kxy018}{doi:\nolinkurl{10.1093/biostatistics/kxy018}}


\bibitem[Sokolova and Lapalme(2009)]%
        {sokolova2009systematic}
\bibfield{author}{\bibinfo{person}{Marina Sokolova} {and} \bibinfo{person}{Guy Lapalme}.} \bibinfo{year}{2009}\natexlab{}.
\newblock \showarticletitle{A systematic analysis of performance measures for classification tasks}.
\newblock \bibinfo{journal}{\emph{Information Processing \& Management}} \bibinfo{volume}{45}, \bibinfo{number}{4} (\bibinfo{year}{2009}), \bibinfo{pages}{427--437}.
\newblock
\href{https://doi.org/10.1016/j.ipm.2009.03.002}{doi:\nolinkurl{10.1016/j.ipm.2009.03.002}}


\bibitem[Thompson et~al\mbox{.}(2017)]%
        {thompson2017communal}
\bibfield{author}{\bibinfo{person}{Luke~R. Thompson}, \bibinfo{person}{Jon~G. Sanders}, \bibinfo{person}{Daniel McDonald}, \bibinfo{person}{Amnon Amir}, \bibinfo{person}{Joshua Ladau}, \bibinfo{person}{Kenneth~J. Locey}, \bibinfo{person}{Robert~J. Prill}, \bibinfo{person}{Anupriya Tripathi}, \bibinfo{person}{Sean~M. Gibbons}, \bibinfo{person}{Gail Ackermann}, \bibinfo{person}{Jose~A. Navas-Molina}, \bibinfo{person}{Stefan Janssen}, \bibinfo{person}{Evguenia Kopylova}, \bibinfo{person}{Yoshiki Vázquez-Baeza}, \bibinfo{person}{Antonio González}, \bibinfo{person}{James~T. Morton}, \bibinfo{person}{Siavash Mirarab}, \bibinfo{person}{Zhenjiang~Zech Xu}, \bibinfo{person}{Lingjing Jiang}, \bibinfo{person}{Mohamed~F. Haroon}, \bibinfo{person}{Jad Kanbar}, \bibinfo{person}{Qiyun Zhu}, \bibinfo{person}{Se~Jin Song}, \bibinfo{person}{Tomasz Kosciolek}, \bibinfo{person}{Nicholas~A. Bokulich}, \bibinfo{person}{Joshua Lefler}, \bibinfo{person}{Colin~J. Brislawn}, \bibinfo{person}{Gregory Humphrey},
  \bibinfo{person}{Sarah~M. Owens}, \bibinfo{person}{Jarrad Hampton-Marcell}, \bibinfo{person}{Donna Berg-Lyons}, \bibinfo{person}{Valerie McKenzie}, \bibinfo{person}{Noah Fierer}, \bibinfo{person}{Jed~A. Fuhrman}, \bibinfo{person}{Aaron Clauset}, \bibinfo{person}{Rick~L. Stevens}, \bibinfo{person}{Ashley Shade}, \bibinfo{person}{Katherine~S. Pollard}, \bibinfo{person}{Kelly~D. Goodwin}, \bibinfo{person}{Janet~K. Jansson}, \bibinfo{person}{Jack~A. Gilbert}, \bibinfo{person}{Rob Knight}, {and} \bibinfo{person}{The Earth Microbiome~Project Consortium}.} \bibinfo{year}{2017}\natexlab{}.
\newblock \showarticletitle{A communal catalogue reveals Earth’s multiscale microbial diversity}.
\newblock \bibinfo{journal}{\emph{Nature}}  \bibinfo{volume}{551} (\bibinfo{year}{2017}), \bibinfo{pages}{457--463}.
\newblock
\href{https://doi.org/10.1038/nature24621}{doi:\nolinkurl{10.1038/nature24621}}


\bibitem[van~der Maaten and Hinton(2008)]%
        {van2008visualizing}
\bibfield{author}{\bibinfo{person}{Laurens van~der Maaten} {and} \bibinfo{person}{Geoffrey Hinton}.} \bibinfo{year}{2008}\natexlab{}.
\newblock \showarticletitle{Visualizing Data using t-SNE}.
\newblock \bibinfo{journal}{\emph{Journal of Machine Learning Research}}  \bibinfo{volume}{9} (\bibinfo{year}{2008}), \bibinfo{pages}{2579--2605}.
\newblock
\newblock
\shownote{Submitted 5/08; Revised 9/08; Published 11/08}.


\bibitem[van Rijsbergen(1979)]%
        {vanrijsbergen1979information}
\bibfield{author}{\bibinfo{person}{C.~J. van Rijsbergen}.} \bibinfo{year}{1979}\natexlab{}.
\newblock \bibinfo{booktitle}{\emph{Information Retrieval} (\bibinfo{edition}{2nd} ed.)}.
\newblock \bibinfo{publisher}{Butterworths}, \bibinfo{address}{Waltham, MA}.
\newblock


\bibitem[Vaswani et~al\mbox{.}(2017)]%
        {vaswani2017attention}
\bibfield{author}{\bibinfo{person}{Ashish Vaswani}, \bibinfo{person}{Noam Shazeer}, \bibinfo{person}{Niki Parmar}, \bibinfo{person}{Jakob Uszkoreit}, \bibinfo{person}{Llion Jones}, \bibinfo{person}{Aidan~N. Gomez}, \bibinfo{person}{Łukasz Kaiser}, {and} \bibinfo{person}{Illia Polosukhin}.} \bibinfo{year}{2017}\natexlab{}.
\newblock \showarticletitle{Attention is All You Need}. In \bibinfo{booktitle}{\emph{Proceedings of the 31st International Conference on Neural Information Processing Systems (NeurIPS)}}. \bibinfo{pages}{6000--6010}.
\newblock
\urldef\tempurl%
\url{https://papers.nips.cc/paper_files/paper/2017/file/3f5ee243547dee91fbd053c1c4a845aa-Paper.pdf}
\showURL{%
\tempurl}


\bibitem[Wold et~al\mbox{.}(1987)]%
        {wold1987principal}
\bibfield{author}{\bibinfo{person}{Svante Wold}, \bibinfo{person}{Kim Esbensen}, {and} \bibinfo{person}{Paul Geladi}.} \bibinfo{year}{1987}\natexlab{}.
\newblock \showarticletitle{Principal Component Analysis}.
\newblock \bibinfo{journal}{\emph{Chemometrics and Intelligent Laboratory Systems}} \bibinfo{volume}{2}, \bibinfo{number}{1--3} (\bibinfo{year}{1987}), \bibinfo{pages}{37--52}.
\newblock
\href{https://doi.org/10.1016/0169-7439(87)80084-9}{doi:\nolinkurl{10.1016/0169-7439(87)80084-9}}


\bibitem[Woloszynek et~al\mbox{.}(2019)]%
        {10.1371/journal.pcbi.1006721}
\bibfield{author}{\bibinfo{person}{Stephen Woloszynek}, \bibinfo{person}{Zhengqiao Zhao}, \bibinfo{person}{Jian Chen}, {and} \bibinfo{person}{Gail~L. Rosen}.} \bibinfo{year}{2019}\natexlab{}.
\newblock \showarticletitle{16S rRNA sequence embeddings: Meaningful numeric feature representations of nucleotide sequences that are convenient for downstream analyses}.
\newblock \bibinfo{journal}{\emph{PLOS Computational Biology}} \bibinfo{volume}{15}, \bibinfo{number}{2} (\bibinfo{date}{02} \bibinfo{year}{2019}), \bibinfo{pages}{1--25}.
\newblock
\href{https://doi.org/10.1371/journal.pcbi.1006721}{doi:\nolinkurl{10.1371/journal.pcbi.1006721}}


\bibitem[Zaheer et~al\mbox{.}(2017)]%
        {zaheer2017deep}
\bibfield{author}{\bibinfo{person}{Manzil Zaheer}, \bibinfo{person}{Satwik Kottur}, \bibinfo{person}{Siamak Ravanbakhsh}, \bibinfo{person}{Barnabas Poczos}, \bibinfo{person}{Ruslan Salakhutdinov}, {and} \bibinfo{person}{Alexander Smola}.} \bibinfo{year}{2017}\natexlab{}.
\newblock \showarticletitle{Deep Sets}. In \bibinfo{booktitle}{\emph{Proceedings of the 31st International Conference on Neural Information Processing Systems (NeurIPS)}}. \bibinfo{pages}{3391--3401}.
\newblock
\urldef\tempurl%
\url{https://proceedings.neurips.cc/paper_files/paper/2017/file/f22e4747da1aa27e363d86d40ff442fe-Paper.pdf}
\showURL{%
\tempurl}


\bibitem[Zhao et~al\mbox{.}(2021)]%
        {10.1371/journal.pcbi.1009345}
\bibfield{author}{\bibinfo{person}{Zhengqiao Zhao}, \bibinfo{person}{Stephen Woloszynek}, \bibinfo{person}{Felix Agbavor}, \bibinfo{person}{Joshua~Chang Mell}, \bibinfo{person}{Bahrad~A. Sokhansanj}, {and} \bibinfo{person}{Gail~L. Rosen}.} \bibinfo{year}{2021}\natexlab{}.
\newblock \showarticletitle{Learning, visualizing and exploring 16S rRNA structure using an attention-based deep neural network}.
\newblock \bibinfo{journal}{\emph{PLOS Computational Biology}} \bibinfo{volume}{17}, \bibinfo{number}{9} (\bibinfo{date}{09} \bibinfo{year}{2021}), \bibinfo{pages}{1--36}.
\newblock
\href{https://doi.org/10.1371/journal.pcbi.1009345}{doi:\nolinkurl{10.1371/journal.pcbi.1009345}}


\bibitem[Zhou et~al\mbox{.}(2023)]%
        {zhou2023dnabert}
\bibfield{author}{\bibinfo{person}{Zhihan Zhou}, \bibinfo{person}{Yanrong Ji}, \bibinfo{person}{Weijian Li}, \bibinfo{person}{Pratik Dutta}, \bibinfo{person}{Ramana Davuluri}, {and} \bibinfo{person}{Han Liu}.} \bibinfo{year}{2023}\natexlab{}.
\newblock \showarticletitle{{DNABERT-2}: Efficient Foundation Model and Benchmark For Multi-Species Genome}.
\newblock \bibinfo{journal}{\emph{arXiv preprint arXiv:2306.15006}} (\bibinfo{year}{2023}).
\newblock
\href{https://doi.org/10.48550/arXiv.2306.15006}{doi:\nolinkurl{10.48550/arXiv.2306.15006}}


\end{thebibliography}





\end{document}